\documentclass{article}

\PassOptionsToPackage{numbers, compress}{natbib}
\usepackage[final]{nips_2017}

\usepackage[utf8]{inputenc} 
\usepackage[T1]{fontenc}    
\usepackage{hyperref}       
\usepackage{url}            
\usepackage{booktabs}       
\usepackage{amsfonts}       
\usepackage{nicefrac}       
\usepackage{microtype}      

\usepackage{multirow}

\usepackage{subfig}
\usepackage{graphicx}

\usepackage{fancyhdr}
\title{Formatting instructions for NIPS 2017}

\begin{document}
\title{Sparse One-Time Grab Sampling of Inliers}
\author{Maryam Jaberi, Marianna Pensky, Hassan Foroosh}



\maketitle

%

\vspace*{-0.2 cm}
Estimating structures in "big data" and clustering them are among the most fundamental problems in computer vision, pattern recognition, data mining, and many other other research fields. Over the past few decades, many studies have been conducted focusing on different aspects of these problems. One of the main approaches that is explored in the literature to tackle the problems of size and dimensionality is sampling subsets of the data in order to estimate the characteristics of the whole population, e.g. estimating the underlying clusters or structures in the data. In this paper, we propose a ``one-time-grab'' sampling algorithm\cite{jaberi2015swift,jaberi2018sparse}. This method can be used as the front end to any supervised or unsupervised clustering method. Rather than focusing on the strategy of maximizing the probability of sampling inliers, our goal is to minimize the number of samples needed to instantiate all underlying model instances. More specifically, our goal is to answer the following question:
{\em ``Given a very large population of points with $C$ embedded structures and gross outliers, what is the minimum number of points $r$ to be selected randomly in one grab in order to make sure with probability $P$ that at least $\varepsilon$ points are selected on each structure, where $\varepsilon$ is the number of degrees of freedom of each structure.''}
This problem can be modeled using hypergeometric pmf. In this paper, we study this model and show the accuracy of each of  the method in choosing the sample size $r$. The steps of the proposed method are summarized as follows:

(i) Estimate probability of selecting zero points in one structure.  
$P_0 \le \left( 1-\frac{r}{C \theta} \right)^\theta\le e^{-r/C}$

\vspace*{-0.05 cm}

(ii) Estimate probability of selecting $\le$ $\varepsilon$ points in one structure
$
\Delta \leq P_0\times \sum_{k=0}^{\varepsilon-1}
{Cd  \choose k}
\left(\frac{\theta}{N - r - \theta + k }\right)^k
$

\vspace*{-0.05 cm}

(iii) Find the upper bounds for the tail probabilities of the multivariate hypergeometric distribution.
$P(\cap_{i=1}^C (d_i\ge \varepsilon)) \geq   1-C\Delta$

Using the non-decreasing property of the above equation, the sample size r can be computed using a binary search. Once sample size $r$ is estimated, a subset of points are sampled uniformly in a single one-time grab to instantiate and cluster the structures in the data.
To verify this prediction, we investigated the accuracy of our approximation $r$ against theoretical values. We chose different population sizes with different embedded model instances. The result of theoretical and estimated values of $r$ are plotted against different desired probability values in the following figure \ref{fig:est-r}. These plots illustrate the average values of $r$ over $200$ independent trials for population sizes of $N=\{100, 1000, 10000\}$. Figure \ref{comp-stat-art} illustrates the sparsity of the proposed method comparing with state-of-the-art methods.\cite{hoseinnezhad2014multi}

\vspace*{-0.5 cm}

\begin{figure}[h]
\centering
\subfloat[Comparison of estimated $r$  versus the theoretical value of $r$, $\varepsilon=2$.]{\includegraphics[trim = 0mm 0mm 0mm 1mm, clip,height=23mm]{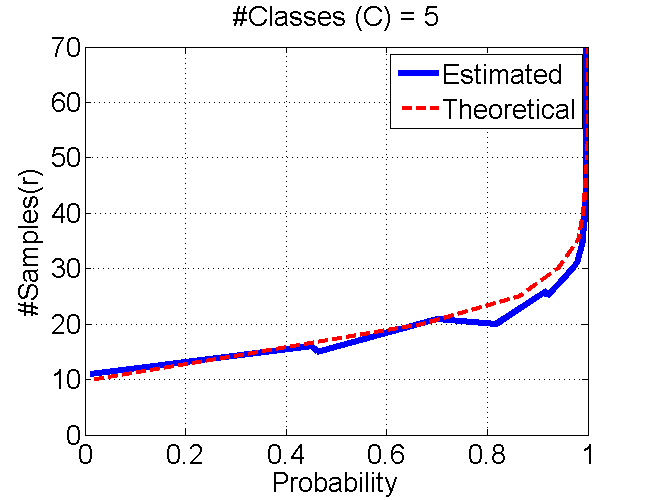}
\includegraphics[trim = 0mm 0mm 0mm 1mm, clip,height=23mm]{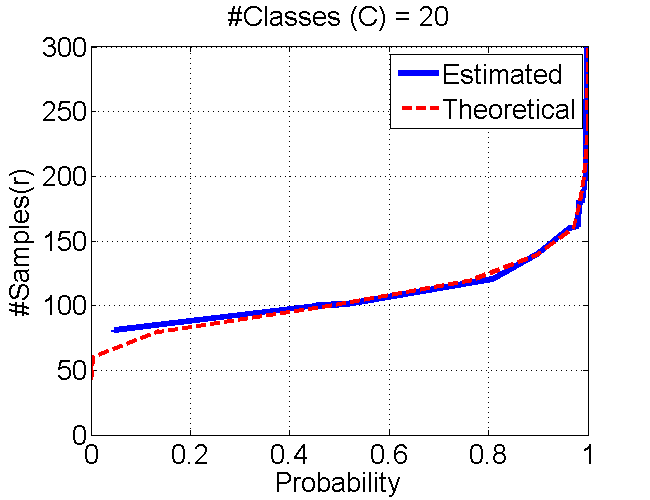}
\includegraphics[trim = 0mm 0mm 0mm 1mm, clip,height=23mm]{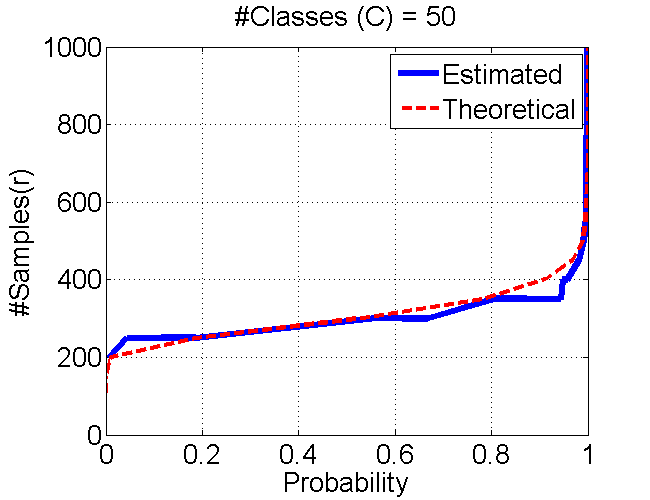}\label{fig:est-r}}\hspace{1mm}
\subfloat[Comparing samples $r$ with state of the arts ]{\includegraphics[trim = 0mm 0mm 0mm 1mm, clip,height=23mm]{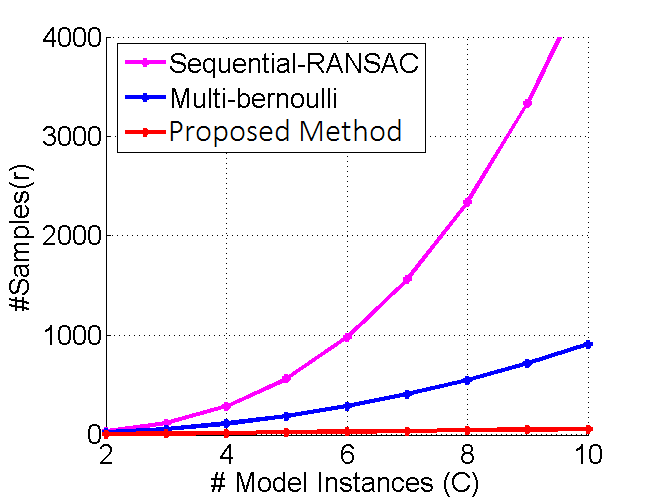}\label{comp-stat-art}}\hspace{1mm}
\captionsetup{labelformat=empty}
\caption{}
\label{fig:r-population}
\end{figure}

\vspace*{-0.9 cm}

As a generic unsupervised sparse sampling method, the proposed sampling method can be used in virtually any scenario where multiple structures need to be detected in a large population of points. Here, a population could be in a physical space (e.g. planar or 3D structures), or in some abstract feature space (e.g. the space of all fundamental matrices, all homographies in some configuration of scene/camera motion, or subspaces formed in some high dimensional spaces\cite{jaberi2018sampling,jaberi2018probabilistic}). Below, we give some examples.

\vspace*{-0.05 cm}

\begin{figure}[h]
\centering
\subfloat[Detecting 3D planes ]{\includegraphics[trim = 6mm 6mm 6mm 5mm, clip, height=23mm]{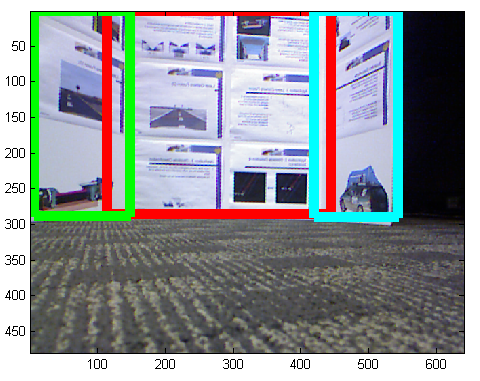} \label{fig:2planes}} \hspace{1mm}
\subfloat[Multibody structure from motion segmentation]
{\includegraphics[trim = 5mm 10mm 10mm 10mm, clip,height=23mm]{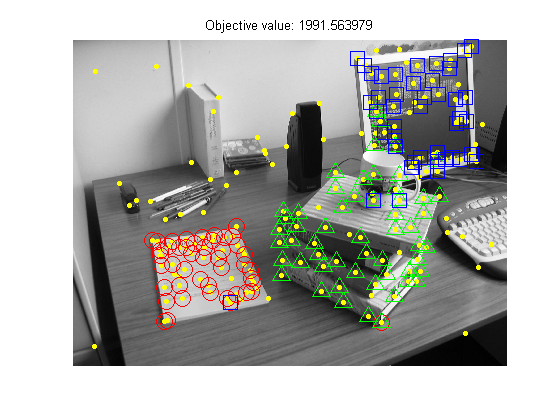}
\includegraphics[trim = 5mm 10mm 10mm 10mm, clip,height=23mm]{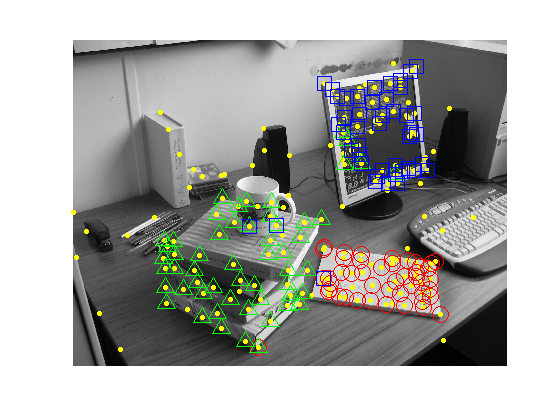}
\label{fig:deskImg:c}}\hspace{0mm}
\subfloat[Face Clustering]{\includegraphics[trim = 0mm 1mm 1mm 1mm, clip, height=23mm]{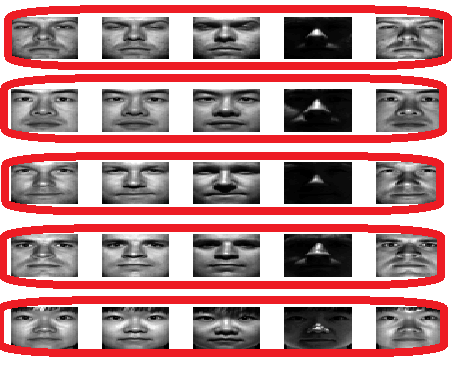} \label{fig:plans_kinect}}
\end{figure}

%


{\small
\bibliographystyle{ieee}
\bibliography{nips_2017}
}

\end{document}